# HETEROGENEOUS KNOWLEDGE REPRESENTATION USING A FINITE AUTOMATON AND FIRST ORDER LOGIC: A CASE STUDY IN ELECTROMYOGRAPHY

Vincent RIALLE

*Faculté de Médecine de Grenoble,* Département de Mathématiques, Statistiques et Informatique Médicale, F-38706 La Tronche Cedex, France.

Annick VILA and Yves BESNARD

*Centre Hospitalier et Universitaire de Grenoble,* Laboratoire d'Electromyographie, Service EFSN, B.P. 217 X, F-38043 Grenoble Cedex, France.

**Abstract**. In a certain number of situations, human cognitive functioning is difficult to represent with classical artificial intelligence structures. Such a difficulty arises in the polyneuropathy diagnosis which is based on the spatial distribution, along the nerve fibres, of lesions, together with the synthesis of several partial diagnoses. Faced with this problem while building up an expert system (NEUROP), we developed a heterogeneous knowledge representation associating a finite automaton with first order logic. A number of knowledge representation problems raised by the electromyography test features are examined in this study and the expert system architecture allowing such a knowledge modeling are laid out.

**Keywords**: Medical expert systems, Heterogeneous knowledge representation, Finite automata, Electromyography.

## 1. Introduction

The various kinds of knowledge and reasoning used in expert systems (ES) have been carefully analyzed and classified over several years [6,11,17]. Nevertheless some types of knowledge remain difficult to represent by means of classical structures (production rules, frames, semantic nets, etc.) commonly used in expert systems. We were faced with this kind of problem while building up an expert system for the electrophysiological diagnosis of neuropathies (NP). The first version of knowledge about NP included in our system was described in 1987 [20].

Several projects have already been completed in the field of artificial intelligence applied to electromyography (EMG). The MUNIN system [2,3] is one of the major realizations. This project is based on a neo-bayesian approach of the inferential process. The reasoning framework is constituted with a probabilistic causal network in which each node has a concurrent conditional probability table. The main task of the system is to calculate probabilities of terminal nodes (illnesses) when some test result values are fed into the network.

The PC-KANDID system was developed in Prolog by A.Fuglsang-Frederiksen's team [7,13]. The system is founded on a logical rule based inference, without uncertainty management, and on an interactive cycle: planning - test - diagnosis. It covers most EMG diagnoses. NEUREX [12], an application of the Parsimonious Covering Theory, is intended to be a general framework for neurological localization and diagnosis. A first outline of an expert system designed for electromyography: MYOSYS, in Prolog language, was outlined in our laboratory by D. Ziebelin [19] in 1984. In contrast to the former systems, which deal with wide fields, the ADELE system [4] is an expert system devoted to the diagnosis of a single disease: the carpal tunnel syndrome. The MYOLOG system [8] realized in Prolog is intended for the diagnosis of cervical radiculopathies and branchial plexus neuropathies, from spontaneous and voluntary muscular activities and clinical data. A system resulting from a neural network and an augmented transition network has been developed in Lisp by P.P. Jamieson [10]. Finally, we should mention the increasing use of artificial neural nets to classify patients into predetermined groups of pathologies [15].

We do not intend to present a new EMG expert system. Our goal is to expose an artificial intelligence modeling problem in a particular field of medical knowledge and practice. The choice of concepts described in this study is based on 2 major reasoning features that lead to NP diagnosis: this diagnosis is based, on one hand, on spatial distribution of lesions along the nerve fibres and, on the other hand, on synthesis of the different local diagnoses (at the nerve segment level, then at the whole nerve level).

Analysis of the lesions distribution along the nerves involves a spatial reasoning that may be either systematic or guided by heuristics. The highlighting of such heuristics, which is part of cognitive psychology [1], seems obviously more complex than the use of systematic reasoning. Nevertheless, the latter, even if described in propositional logic, would still be cumbersome with regards to the considered reasoning features. Therefore, the knowledge modeling solution we propose is heterogeneous: it is based on a finite automaton [9,14] used jointly with three other knowledge sources. First order logic is used in almost every part of the system : inference engine, finite automaton and production rules. The overall system is written in Prolog [5,16,18].

First we will briefly describe the main characteristics of an EMG examination. Then, we will lay out the general structure of the EMG knowledge leading to the diagnosis of neuropathies. A number of knowledge representation problems raised by the EMG test characteristics will be tackled and the choice of a heterogeneous knowledge representation using a finite automaton, as a solution to these problems, will be explained. Due to the number of knowledge sources, we will not describe them extensively but we will focus in greater detail on the description of the automaton and on the overall expert system architecture which make its use possible.

**2. EMG principles**

EMG consists of different suitable electrophysiological techniques aimed at studying the peripheral nerve system (PNS), muscles and neuromuscular junction. Neuropathies are diseases of the PNS and especially of the nerves innervating the face and limb muscles. Lesions detected by the EMG examination affect either axons of the fibres (*axonal lesions*) or myelin sheath (*demyelinating lesions*); if both structures are concerned, we call them *mixed lesions*. The neuropathy diagnosis is provided by the EMG examination of several nerves, mainly median and ulnar nerves for the upper limbs, peroneal and tibial nerves for the lower limbs.

The diagnosis activity involves 2 main steps, with a specific type of knowledge for each step. These two steps are as follows:

• *Step A*: successive tests are performed on each chosen nerve. The choice of nerves is made according to the patient's symptoms and signs. These tests are included into charts that have been standardized in our laboratory. They are chosen according to the clinical findings — i.e. the diagnosis hypotheses —, and to the nerve under study. The number of segments that we usually explore is variable: generally between 1 and 2 for sensitive fibres, and between 2 and 5 for motor fibres. For example, in the median nerve study, there are 2 stimulation points — palm and wrist — for the sensory nerve conduction test, and 5 segments for the motor nerve conduction test.
Electrophysiological parameters to be analyzed on the sensory and muscle action potentials are:
a) sensory fibres: sensory action potential (SAP) and sensory nerve conduction velocity (SCV) on both segments; amplitude ratio of the SAP for the second segment.
b) motor fibres:
  - for the first segment: amplitude and distal latency of the compound muscle action potential (CMAP),
  - for the next four segments: amplitude of the CMAP, amplitude ratio of the successive CMAP, and motor nerve conduction velocity (MCV).
Each such examined nerve is given a diagnosis, whose process will be described below.

• *Step B*: the overall diagnosis is made up from the synthesis of the diagnoses for each nerve (*local* synthesis) and for the various nerves (*overall* synthesis).

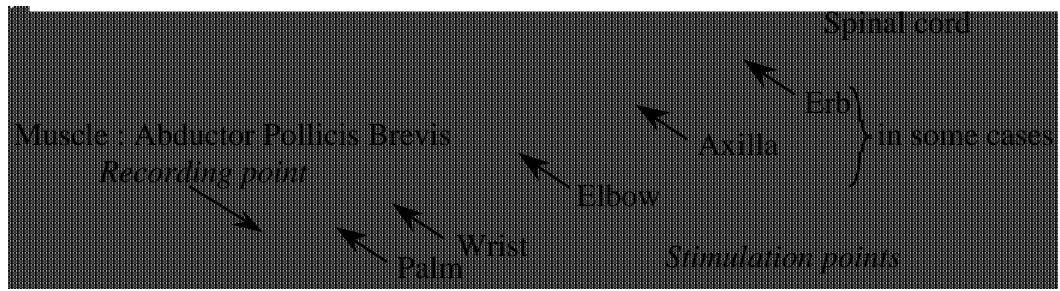

Fig. 1. Motor median nerve segmentation

## 3. Knowledge description and representation

*3.1. knowledge levels*

These two stages of the examination lead to the three following levels of knowledge (Fig. 2):

*Level 1* : analysis of one nerve segment (step A,1),
*Level 2* : analysis of a total nerve (step A,2): synthesis of the segment analyses,
*Level 3* : analysis of the whole nerves (step B): synthesis of the nerve analyses.

Thus, both level 2 and level 3 correspond to the synthesis of the preceding level.

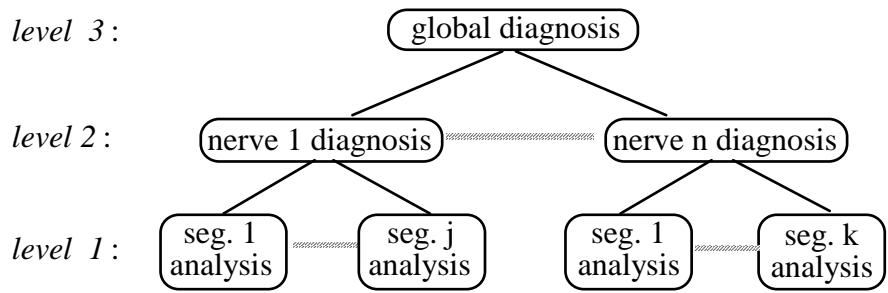

Fig. 2. Diagnosis analysis tree

*3.2. Knowledge representation at different levels*

*Level 1*: there are three types of segments depending, on the one hand, on the kind of fibre under consideration (sensory or motor), on the other hand, on the position of the segment with regards to the detection point. For a given type of segment, the diagnosis is given by about ten production rules. All electrophysiological variables, such as amplitudes, velocities, amplitude ratios, etc., are continuous. Their values are interpreted by an interpretation module. This module creates the corresponding semantic variables getting their values in sets such as {normal, decreased, very decreased} or {normal, increased, very increased}. These semantic variables are used as premises to the segment diagnosis rules (Fig. 3).

|     |     |                                                              |     |                    |
| --- | --- | ------------------------------------------------------------ | --- | ------------------ |
|     | *if*  | amplitude at the wrist                                     | is  | very decreased     |
| *and* | *if*  | ratio: amplitude at the wrist / amplitude at the palm      | is  | normal             |
| *and* | *if*  | velocity at the wrist                                      | is  | normal or decreased |
|     | *then* | lesion of the segment                                     | is  | **severe axonal**  |
|     | *if*  | amplitude at the wrist                                     | is  | normal             |
| *and* | *if*  | ratio: amplitude at the wrist / amplitude at the palm      | is  | normal or decreased |
| *and* | *if*  | velocity at the wrist                                      | is  | decreased          |
|     | *then* | lesion of the segment                                     | is  | **mild demyelinating** |

Fig. 3. Sample of diagnosis rules applied to a motor segment of the median nerve

*Level 2*: Each segment whose diagnosis is different from the *normal* label is *pathological* (we do not take into consideration, at this stage, the kind of lesion: axonal, demyelinating or mixed). The knowledge leading to the diagnosis of the nerve is based on the spatial distribution of the pathological segments along the nerve (Fig. 4).
In level 2 there are three possible kinds of pathologies, determined according to the following rules:

- *focal neuropathy:* only one segment is affected
- *multiple focal neuropathy:* several non-contiguous segments are affected
- *diffuse neuropathy:* several contiguous segments are affected

```
  0    1    0    0    0     focal neuropathy
  0    1    0    1    0     multiple focal neuropathy
  1    0    1    1    0     diffuse neuropathy
```
(0: normal ; 1: lesion)

Fig. 4. Examples of spatial distribution of lesions

*Level 3*: As in level 2, knowledge leading to the diagnosis of all the nerves is based on a spatial distribution of the nerve lesions as well as on the number of lesions, with moreover a notion of symmetry (Fig. 5) (symmetrical nerves are called *homologous* ).

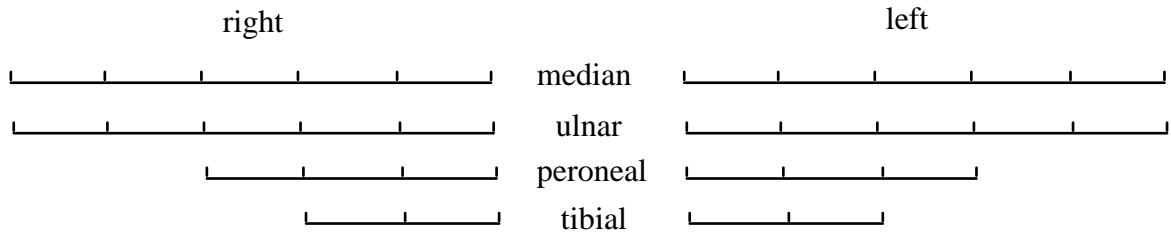

Fig. 5. Diagram of spatial distribution of the segments

The possible overall diagnoses are the following: { focal mono-neuropathy, multiple focal neuropathy, diffuse mono-neuropathy, symmetrical poly-neuropathy, asymmetrical poly-neuropathy, uncertain diagnosis, normal examination }. The processing rules of association between the affected nerves and the diagnosis are as follows:

- focal mono-neuropathy:           one effected segment (a.s.) on only one nerve
- multiple focal mono-neuropathy: several non-contiguous a.s.
                                   (distributed or not on the whole nerves)
- diffuse mono-neuropathy:         at least two contiguous a.s. on the same nerve
- symmetrical poly-neuropathy:     at least two contiguous a.s. on homologous nerves
- asymmetrical poly-neuropathy:    at least two contiguous a.s. on non-homologous nerves
- normal examination:              no a.s.
- uncertain diagnosis:             any other case

**4. Problems of knowledge representation at level 2**

Each of the three knowledge levels that have just been described needs its own model of representation. Level 1 corresponds to a simple classification knowledge that directly associates, for any segment, the electrophysiological values to the diagnosis. This knowledge does not pose any problem of representation.

Level 2 is where a representation problem starts. Indeed, as previously stated, the diagnosis of the whole nerve needs a particular element of reasoning: the *relative location of lesions.*. The practitioner's appreciation of this location stems from an empirical knowledge. The distribution of the segments affected by a lesion could be represented by a categorical variable, called *relative location of the lesions,* the value of which is in the set : {unique isolated lesion, non-contiguous isolated lesions, contiguous lesions}. The associated diagnosis is obtained by application of rules described in point 3.

The most commonplace solution would consist in creating such a variable and proposing the following choice to the user (by means of a dialogue screen and whenever the reasoning needs it): a) no segment affected, b) a unique segment affected, c) several isolated segments affected, d) several contiguous segments affected. Obviously, the consequence of this solution would be to considerably diminish the degree of *intelligence* of the system, since it is possible to deduce the answer from the state of the factual base.

A second solution would consist in representing all the possible figures in production rules such as: *if* $seg_1$ is $V_1$ and $seg_2$ is $V_2$ and ... and $seg_n$ is $V_n$ *then* $nerve_k$ is $W_k$; with $V_i$ { normal, pathological } and $W_k$ { normal examination, focal neuropathy, multiple focal neuropathy, diffuse neuropathy }.

Nevertheless, this solution has two major drawbacks:
- *First*: the number of possibilities for one nerve with n segments is $2^n$. Since, first, n is comprised between 1 and 5 and, second, there are at least 20 couples of rachidian nerves likely to be analyzed, we obtain an explosion of potential combinations, even in first order logic.
- *Second*: from a scientific or medical point of view these production rules have no interest since they are simple basic operation rules.

**5. Finite automaton**

Taking into account these knowledge representation problems, we made the choice of inserting in the knowledge base and in the inference engine an extremely simple and safe mechanism with regards to its principle as well as its recursive representation in first order logic: the *finite automaton*. This solution allows an almost immediate association between the list of segment statements and the nerve diagnosis.

*5.1. Definition*
A finite automaton consists of a finite set of states and a set of transitions which make it possible to move from one state to another when a symbol, from a predetermined set of input symbols, is fed into the automaton. The finite automaton is formally defined by a 5-tuplet $(Q, \Sigma, \delta, q_0, F)$ in which Q is the finite set of states, $\Sigma$ is a finite input alphabet, $q_0$ belongs to Q and designates the automaton initial state, $F \subseteq Q$ is the set of final states, and $\delta$ is the transition function associating each couple (q,a) of Q x $\Sigma$ with an element p' of Q: $\delta(q,a) = q'$.

With regards to our application, the finite set of states is Q = { start, n, f_a, f_b, m_f_a, m_f_b, d }. The symbols n, f_a, m_f_a and d represent normal, focal, multiple focal and diffuse states respectively. These states are represented by large triangles in the transition diagram (Fig. 6). The symbols f_b and m_f_b (small triangles in the diagram) correspond to states which repeat a principal state and yet direct the analysis towards a new principal state.

The set of final states is: F = {n, f_a, f_b, m_f_a, m_f_b, d}.

As indicated in 3.2, the normal segments are symbolized by 0 and the affected ones by 1, hence: $\Sigma$ = { 0, 1 }.

For example, the following input chain : [0,1,0,1,0], called *chain of segment states*, reflects the state of a nerve whose second and fourth segments are affected, the three others being normal. Therefore the definition of an automaton for the analysis of a nerve is : (Q, $\Sigma$, $\delta$, start, F).

The symbolic functioning is the following: the chain of segment states is read from left to right or from right to left (the direction is not important). Each reading of a new value from the input chain introduces a *state transition* from the previous state. For example, let n be the automaton state at a given time of the analyzing process, and let 1 be the new input value, then the automaton moves to the state f_a (i.e. $\delta(n,1)$ = f_a). The initial state, before the first input, is the *start* state. The final state (after the last input) corresponds to the nerve diagnosis. Since there are several transition possibilities (exactly 2) for every state, the automaton is called *nondeterministic*.

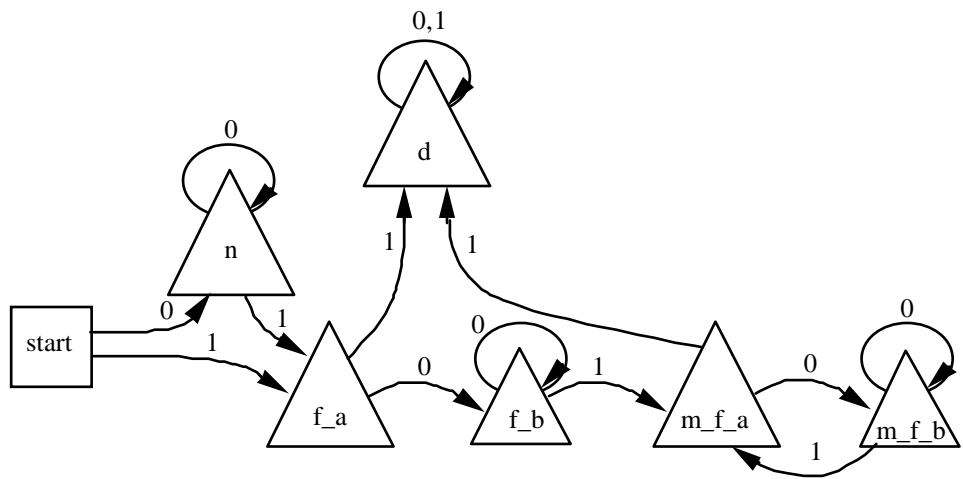

Fig. 6.   Transition diagram of the finite automaton

n : normal state          f_a, f_b :     focal neuropathy

d : diffuse neuropathy    m_f_a, m_f_b : multiple focal neuropathy

The transitions defining the automaton and expressed in relational form, as in Prolog language, are shown in Figure 7. The automaton mechanism is expressed in the recursive predicate shown in Figure 8.

```
transition ( start , 0 , n ).
transition ( start , 1 , f_a ).
```

```
transition ( n , 0 , n ).
transition ( n , 1 , f_a  ).
transition ( f_a , 0 , f_b ).
transition ( f_a , 1 , d ).
transition ( f_b , 0 , f_b ).
transition ( f_b , 1 , m_f_a ).
transition ( m_f_a , 0 , m_f_b ).
transition ( m_f_a , 1 , d ).
transition ( m_f_b , m_f_b , 0 ).
transition ( m_f_b , 1 , m_f_a ).
transition ( d , 0 , d ).
transition ( d , 1 , d  ).
```

Generical form : transition (<former state>,<current input>,<resulting state>)

Fig. 7. Automaton transition relations

```
automaton ([], Final_state, Final_state).
automaton ([_], d, d).
automaton ([T|Q], Previous_state, Final_state):-
     transition ( Previous_state, T, New_state),
     automaton (Q, New_state,Final_state),!.
```

Fig. 8. Finite automaton predicate

### 5.2. Utilization

The finite automaton described above simulates the knowledge of level 2. One can consider that it is in itself a small-scale system (inference engine + knowledge base) in which the motor is constituted by the recursive predicate *automaton*, and the knowledge base is represented by the transition set. Thus, the automaton insertion in the expert system takes place at the inference engine level as well as at the knowledge base level.

The system architecture includes 4 cooperating knowledge sources successively used through the 4 phases of the diagnosis process (Fig. 9). A special inference engine supervises the successive use of the knowledge bases. These knowledge bases communicate via a central working memory, as in the case of a *blackboard*.

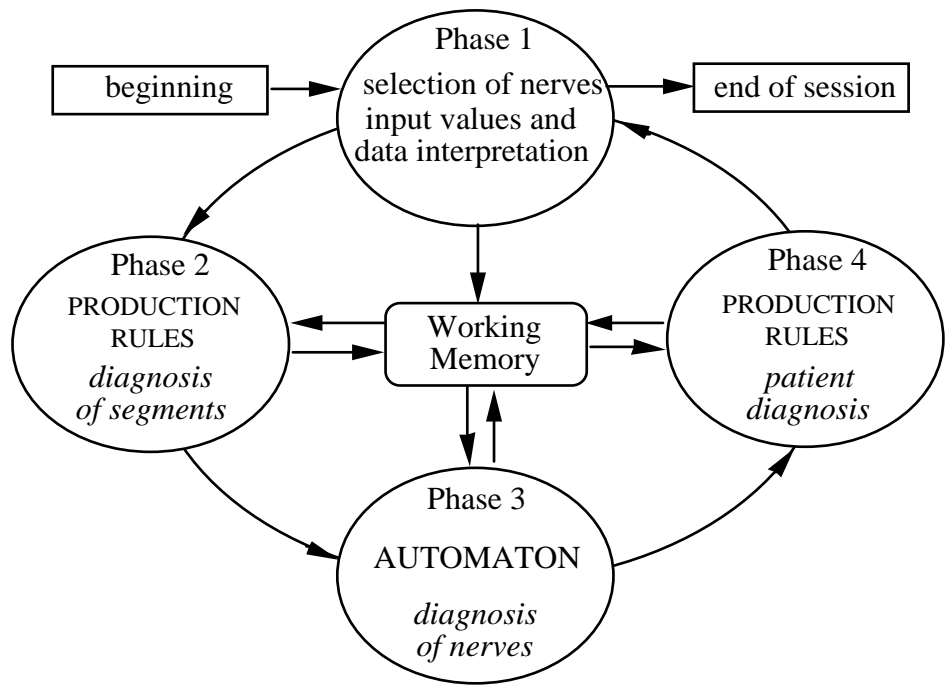

Fig. 9. Location of the automaton in the diagnosis research process

The 4 temporal phases of the process are the following:

- phase 1: nerve selection by the physician, input and interpretation of electrophysiological parameters,
- phase 2: inference cycles on each segment (segment diagnosis),
- phase 3: automaton activation on each nerve (nerve diagnosis),
- phase 4: inference cycles on all nerves (patient diagnosis),

The *working memory* allows result sharing between the different inference phases, i.e.:
- from phase 1 to phase 2: sharing of the electrophysiological parameters,
- from phase 2 to phase 3: sharing of the diagnoses of segments,
- from phase 3 to phase 4: sharing of the diagnoses of nerves.

**6. Conclusion**

We showed that there are some scientific knowledge fields for which classical structures such as production rules, frames or semantic nets are not suitable. EMG is a typical example of such fields: the physician performs a spatial reasoning from the distribution and number of abnormal segments. Such a reasoning is without doubt the result of different procedures depending on the physician (competence, experience, etc.). We leave it to cognitive psychology to study human functioning in establishing such a diagnosis. Thus, the model we described *is not a priori* a human function model. The synthesis diagnosis is set up by a finite automaton which repeats a sequential, spatial and numeric reasoning. This reasoning is rather similar to that of a beginner's or a very well organized but nervous person's.

The automaton stands among three other knowledge sources, whose functioning is supervised by an appropriate inference engine. The insertion of this automaton in the core of the expert system has notably influenced the architecture of the whole system. First order logic is used in the construction of the automaton as well as in the representation of knowledge of levels 1 and 3.

This solution allows a significant decrease of the number of production rules and of the system response time. The current version of the system does not use any probability or uncertainty weighting. A new version using an approximate reasoning is under study.

This heterogeneous knowledge representation is operational in the NEUROP expert system, written in Prolog and currently under evaluation in the EMG Laboratory of the teaching hospital of Grenoble.

**7. Discussion**

The paradigm of expert systems includes the idea that knowledge accumulated in a knowledge base should reflect the expert's knowledge as exactly as possible. Classical knowledge representation structures such as production rules, frames or semantic nets must allow an easy *reading* of this knowledge for maintenance and updating purposes. This legitimate point of view has contributed to the wide development of knowledge based systems. However, such an opinion is limited to knowledge that easily lends itself to a symbolic representation (such a feature does not in any way preclude the intrinsic knowledge complexity).

It is not so when the conceptual model cannot be easily translated into classical structures such as production rules, frames, semantic nets or mixed models combining two or three of these structures. Even though it is not closely copied from these usual artificial intelligence structures, the automaton model is yet connected to the ES paradigm at least through its declarative form in first order logic. Moreover, as indicated in point 5, this model could be considered as a small-scale production system. Particularly, there is a strong analogy between the transition relation set as shown in fig. 7 and a small knowledge base. Especially, this transition set could be easily altered by addition, modification or withdrawal of transitions, according to a possible knowledge improvement. This will in fact be the case in the next version of the automaton, when specialists include knowledge related to *conduction blocks* (null amplitude of a sensory or compound muscle action potential ) in the expert system.

Nevertheless the similitude to the ES paradigm remains limited. Indeed, there is no reason for affirming that the model constituted by the finite automaton reflects an expert's way of reasoning (it is difficult to imagine that an expert reasons as mechanically as the automaton does).

However, one major question raised from the adoption of a procedural knowledge representation in an expert system is: is it necessary to represent, with an absolute accuracy, the expert's way of reasoning? Is this accuracy not achieved in some cases at the cost of a considerable complication of the knowledge base?

The weakness that one could reproach the automaton model with is that it does not use any heuristic knowledge — a major attribute of human intelligence —: the way to achieve the synthesis diagnosis of a nerve is an absolutely systematic and exhaustive reasoning. However, this weakness is compensated by the relatively low number of nerve segments (5 maximum) which minimizes the advantage of heuristic rules.

It should also be observed that, to be understood, the automaton we have devised needs very few concepts about the theory of finite automata, and constitutes an elegant solution to the spatial reasoning knowledge representation. As far as they are concerned, the medical experts who participated in the knowledge base building consider the adoption of the notion of the finite automaton as a practical substitute to a human knowledge model whose conception is beyond their reach. This way, we can state that such a solution constitutes to some extent a contribution of artificial intelligence to the medical expertise modeling.

**Acknowledgements**. This work has been supported by the REGION RHONE-ALPES via the Pôle Rhône-Alpes de Génie Biologique et Médical under grant 509004/12 GBM.

References

[1] J.R. Anderson, Cognitive Psychology, *Artificial Intelligence* 23 (1984) 1-11.

[2] S. Andreassen, S.K. Andersen, F.V. Jensen, M. Woldbye, A. Rosenfalck, B. Falck, U. Kjærulff and A.R. Sorensen, MUNIN - An expert system for EMG, *Electroencephalography and Clinical Neurophysiology* 66 (1987) S4.

[3] S. Andreassen and M. Wellman, MUNIN - On the case for probabilities in medical expert systems - a practical exercise, in: J. Fox, M. Fieschi and R. Engelbrecht (Eds.), *Lecture Notes in Medical Informatics* 33 (Springer Verlag, New York, 1987) 149-160.

[4] A. Blinowska and J. Verroust, Building an expert system in electrodiagnosis of neuromuscular diseases: prototype system, *Electroencephalography and Clinical Neurophysiology* 66 (1987) S10.

[5] I. Brakto, *Prolog Programming for Artificial Intelligence* (Addison-Wesley Publishing, Massachusetts, 1986).

[6] W.J. Clancey, Heuristic Classification, *Artificial Intelligence* 27 (1985) 289-350.

[7] A. Fuglsang-Frederiksen, J. Ronager and S. Vingtof, PC-KANDID: An expert system for electromyography, *Artificial Intelligence in Medicine* 1 (1989) 117-124.

[8] R. Gallardo, M. Gallardo, A. Nodarse, S. Luis, R. Estrada, L. Garcia and O. Padron, Artificial Intelligence in the electromyographic diagnosis of cervical roots and brachial plexus lesions, *Electroencephalography and Clinical Neurophysiology* 66 (1987) S37.

[9] J.E. Hopcroft and J.D. Ullman, *Introduction to automata theory, languages and computation* (Addison-Wesley Publishing, Massachusetts, 1979).

[10] P.W. Jamieson, Symbolic analysis of electromyographic data, *Electroencephalography and Clinical Neurophysiology* 66 (1987) S50.

[11] E.T. Keravnou and J. Washbrook, Deep and Shallow models in medical expert systems, *Artificial Intelligence in Medicine* 1 (1989) 11-28.

[12] J.A. Reggia, S. Tuhrim, S.B. Ahuja, T. Pula, B. Chu, V. Dasigi and J. Lubel, Plausible Reasoning During Neurological Problem Solving: The Maryland NEUREX Project, in: R. Salamon, B. Blum and M. Jorgensen (Eds.), *Proc. MEDINFO 86* (Elsevier Science Publishers, Amsterdam, 1986) 17-21.

[13] J. Ronager, S. Vingtof, J.O. Kjærum and A. Fuglsang-Frederiksen, An EMG expert assistant in clinical neurophysiology (PC-KANDID), *Theoretical Surgery* 3 (1988) 48.

[14] A. Salomaa, *Theory of automata* (Pergamon Press, Oxford, 1969).

[15] C.N. Schizas, C.S. Pattichis, I.S. SChofield, P.R. Fawcett and L.T. Middleton, Artificial neural net algorithms in classifying electromyographic signals, in: *Proc. First IEE International Conference on Artificial Neural Networks* (Institution of Electrical Engineers, London, 1989) 134-138.

[16] E. Shapiro and L. Sterling, *The art of Prolog* (M.I.T. Press, Massachusetts, 1987).

[17] J.F. Sowa, *Conceptual structures: information processing in mind and machine* (Addison-Wesley Publishing, Massachusetts, 1984).

[18] P.A. Subrahmanyam, The software engineering of expert systems: Is Prolog appropriate? *IEEE transactions on software engineering* SE-11 (1985) 11.

[19] A. Vila, D. Ziebelin and F. Reymond, Experimental EMG expert system as an aid in diagnosis, *Electroencephalography and Clinical Neurophysiology* 61 (1985) S240.

[20] A. Vila, L. Kress, V. Rialle, C. Robert and D. Ziebelin, Are expert-systems an aid for diagnosing neuropathies ?, *Electroencephalography and Clinical Neurophysiology* 66 (1987) S109.